# Integrated Sequence Tagging for Medieval Latin Using Deep Representation Learning

**Mike Kestemont[1], Jeroen De Gussem[2]**

1 University of Antwerp, Belgium

2 Ghent University, Belgium

*Corresponding author: mike.kestemont@uantwerpen.be

**Abstract**
In this paper we consider two sequence tagging tasks for medieval Latin: part-of-speech tagging and lemmatization. These are both basic, yet foundational preprocessing steps in applications such as text re-use detection. Nevertheless, they are generally complicated by the considerable orthographic variation which is typical of medieval Latin. In Digital Classics, these tasks are traditionally solved in a (i) cascaded and (ii) lexicon-dependent fashion. For example, a lexicon is used to generate all the potential lemma-tag pairs for a token, and next, a context-aware PoS-tagger is used to select the most appropriate tag-lemma pair. Apart from the problems with out-of-lexicon items, error percolation is a major downside of such approaches. In this paper we explore the possibility to elegantly solve these tasks using a single, *integrated* approach. For this, we make use of a layered neural network architecture from the field of deep representation learning.

**keywords**
tagging, lemmatization, medieval Latin, LSTM, deep learning

## INTRODUCTION

Latin —and its historic variants in particular— have long been a topic of major interest in Natural Language Processing [Piotrowksi 2012]. Especially in the community of Digital Humanities, the automated processing of Latin texts has always been a popular research topic. In a variety of computational applications, such as text re-use detection [Franzini et al, 2015], it is desirable to annotate and augment Latin texts with useful morpho-syntactical or lexical information, such as lemmas. In this paper, we will focus on two sequence tagging tasks for medieval Latin: part-of-speech tagging and lemmatization. Given a piece of Latin text, the task of lemmatization involves assigning each word to a single dictionary headword or 'lemma': a baseform label (preferably in a normalized orthography) grouping all word tokens which only differ in spelling and/or inflection [Knowles et al, 2004]. The task of lemmatization is closely related to that of part-of-speech (PoS) tagging [Jurafsky et al, 2000], in which each word in a running text should be assigned a tag indicating its part of speech or word class (e.g. noun, verb, adjective, etc.). The difficulty of PoS-tagging strongly depends of course on the complexity and granularity of the tagset chosen. Lemmatization and PoS-tagging are classic forms of sequence labeling, in which tags are assigned to words, both on the basis of their individual appearance, as well as the other words which surround them.

## I DATA, STATE OF THE ART

### 1.1. Challenges in Lemmatizing and Tagging Medieval Latin

While both lemmatization and PoS-tagging are rather basic preprocessing steps, they are generally complicated by a number of interesting challenges which the Latin language poses. First of all, while plain stemming might take us a long way [Schinke et al, 1996], many Latin suffixes cannot be automatically linked to an unambiguous morphological category. Words ending in *–ter*, for example, correspond to no less than six different parts of speech: nouns (*fra-ter*), adjectives (*dex-ter*), pronouns (*al-ter*), adverbs (*gravi-ter*), numeral adverbs (*qua-ter*) and prepositions (*in-ter*) [Manuel de lemmatisation, LASLA, 2013]. Additionally, like many other languages, Latin is teeming with homographs which require context to be disambiguated. A token such as *legi* can both be lemmatized under the verb lego as under the noun *lex*. Similarly ambiguous tokens include common forms such *quae*, *satis* or *venis*. For lemmatization specifically, another problem is verb forms which show no resemblance to their lemma. The fact that *tuli* is an 'active 1st person singular perfect' of *fero* is not obvious, and the same problem applies to *fero*'s perfect participle *latus*, which could in its own turn be confused with the homonymous common noun *latus* (transl. "side"). A tagger has to learn the morphological connection between *tuli, latus* and *fero* by moving beyond superficial outward appearances (prefixes, word stems or suffixes), and by properly modelling the immediate context surrounding these words.

Latin, as a school-preserved language, changed surprisingly little throughout its history compared to other languages. Nevertheless, it did witness the introduction of considerable orthographical variation [Rigg 1996], affecting both the spelling and spacing words, especially in medieval times. Some variants in Latin spelling are based on convention and relate to editorial preferences, rather than linguistic evolutions. Well known are the orthographical alterations between <v> and <u> (e.g. *auus*, *avus* versus *avvs*) and <i> and <j> (e.g. *jejunium* and *ieiunium*) to distinguish between vowels and consonants, a post-medieval distinction which occurs only from the 17th-18th century onwards. Earlier phonological evolutions within Latin have caused other orthographical peculiarities. An important example is the evolution from the classical diphtongs <ae> or <oe> to <e> (*aetas* vs *etas*), which in their own turn caused the occurrence of hypercorrected forms such as *aecclesia* instead of *ecclesia*. The implication of the preceding example is that normalizing the spelling of a word is not a simple conversion task which goes in one direction only (because <e> to <ae> is not a rule in se if we sometimes need to correct <ae> to <e>). The ambiguity between <ae> and <e> is problematic, since both can function as case endings and consequently carry relevant inflectional information which one would like the tagger to detect. Other examples of linguistic deviation found in medieval texts are: the alteration between <ti> and <ci> caused by lenition (e.g. *pronuntio* vs *pronuncio*), the alteration between <e> and <i> because of confusion between the long vowel /e:/ and the short /i/ (e.g. *quolebet* vs *quolibet*), the loss or addition of initial <h> (e.g. *habundanter* for *abundanter*, or *ebdomada* for *hebdomada*), strengthened aspiration or fortition (e.g. *michi* for *mihi* or *nichil* for *nihil*), the intrusion of <p> after an <m> (e.g. *hiemps* for *hiems*, or *dampnum* for *damnum*), etc.

Naturally, orthographical artifacts and homography pose challenging problems from a computational perspective [Piotrowski 2012]. Consider the surface form *poetae* to which, already, three different analyses might be applicable: 'nominative masculine plural', 'genitive masculine singular' and 'dative masculine singular'. In medieval texts, the form *poetae* could easily be spelled as *poete*, a spelling which in its turn causes confusion with other declensions, such as *dux*, *duc-e*. Thus, a good model of the local context in which ambiguous word forms appear is crucial to their disambiguation. Nevertheless, Latin is a highly synthetic language which generally lacks a strict word order: it is therefore far from trivial to extract syntactic patterns from Latin sentences (e.g. an adjective modifying a noun does not necessarily immediately proceed or follow it). This lack of a strict word order on the one hand and the concordance of morphological features on the other can cause Latin to display a number of *amphibologies* or so called "crash blossoms". These are sentences which allow for different syntactic readings (e.g. *nautae poetae mensas dant*).[1]

---

[1] This specific example could be translated as "the seafarers give food to the poet", "the poets give food to the seafarer", "the seafarers give the food of the poet", "the seafarers give the food of the poet", etc. Grammatically

## 1.2 Survey of resources and related research

While basic sequence tagging tasks such as PoS-tagging are typically considered 'solved' for many modern languages such as English, these problems remain more challenging for so-called 'resource-scarce' languages, such as Latin, for which fewer or smaller resources are generally available, such as annotated training corpora. In this section, we review some of the main corpora which are currently available, including a short characterization, and a brief description of the type of annotations they include.

### *1.2.1 Index Thomisticus*

From early onwards, Latin was an important research topic in the emerging community of Digital Humanities, more specifically with the undertaking of the *Index Thomisticus* by Roberto Busa, s.j. in the second half of the 1940s [Passarotti 2013]. The corpus contains all 118 texts of 13th century author Thomas Aquinas as well as 61 texts which are related to him, approximating ±11,000,000 words which can be searched online. The website additionally allows to compare and sort words, phrases, quotations, similitudes, correlations and statistical information. In 2006, the *Index Thomisticus* team started a treebank[2] project in close collaboration with the *Latin Dependency Treebank* [Passarotti et al, 2010; 2014]. Their annotation style was inspired by that of the Prague Dependency Treebank and the Latin grammar of Pinkster [Bamman et al, 2007]. The IT-TB training sets, taken from Thomas Aquinas' *Scriptum super Sententiis Magistri Petri Lombardi*, are available for download in the CoNLL-format and comprise over ±175,000 tokens. The *Index Thomisticus*, with its present treebank venture, is a seminal project that until this very day proves to be of considerable value to the progress of Latin automatic annotation.

### *1.2.2 Latin Dependency Treebank*

A second project occupied with Latin treebanking is the *Latin Dependency Treebank* (LDT), which was developed as a part of the Perseus Project at the Tufts University in 2006. Classical texts from Caesar, Cicero, Jerome, Vergil, Ovid, Petronius, Phaedrus, Sallust and Suetonius were manually annotated by adopting the Guidelines for the Syntactic Annotation of Latin Treebanks (cfr. supra), resulting in a corpus of ±53,000 words which was made available online. Treebanking implies full parsing information (syntactic and semantic annotation), whereas for us only the morphological information included in the PoS-tag is relevant.

### *1.2.3 PROIEL*

Another noteworthy treebank project is *PROEIL* (*Pragmatic Resources of Old Indo-European Languages*) [Haug and Jøhndal, 2008]. Its goal is to find information structure systems cross-linguistically over the different translations of the Bible (Latin, Greek, Gothic, Armenian and Church Slavonic). In a first phase, these texts were automatically PoS-tagged and manually corrected. A rule-based 'guesser' consequently suggested the most likely dependency relation for the annotator. Their annotation scheme for syntactic dependencies was based on that of the LDT, but they have fine-grained the domain of verbal arguments and adnominal functions [Haug and Jøhndal, 2008]. This training data has been made available online (in the CoNLL standard), and includes roughly 179,000

---

each of these translations can be considered correct, although it is likely that the best option can either be inferred from context or from the language's patterns of word order (which exhibits itself more often in prose than in poetry) [Devine and Stephens, 2006]. The statement that Latin does not have a structuralized sentential order has to be nuanced.

[2] A syntactically and/or semantically parsed corpus of text.

Latin words from Jerome's *New Testament*, Cicero's *Letters to Atticus*, Caesar's *De Bello Gallico* and the *Peregrinatio Egeriae*.

*1.2.4 LASLA*

A fourth project worth mentioning is *LASLA* (*Laboratoire d'Analyse Statistique des Langues Anciennes*). This project has developed a lemmatized corpus, comprising classical texts such as those of Caesar, Catullus, Horace, Ovid and Virgil, which can be searched online if registered, but is not publicly available for download. Their lemmatization method of Latin is, however, semi-automatic. Firstly, the word is automatically analyzed on the basis of its stem and case ending, which results in a list of possible lemmas. At this stage, the choice of the correct lemma and its correct morphological analysis occurs manually, which is a rather time-consuming undertaking [Mellet and Purnelle, 2002]. For a reference dictionary in producing the lemmas, *LASLA* has used Forcellini's *Lexicon totius latinitatis*, with the reasonable argument that it is the least incoherent [Manuel de lemmatisation, LASLA, 2013].

*1.2.5* (*CHLT*) *LemLat*

CHLT *LemLat* is a Neo-Latin morphological analyzer, the first version of which appeared in 1992. It was "statistically able to lemmatize ±1,300,000 wordforms from the origins to the fifth/sixth century after Christ" [Bozzi et al, 2002]. CHLT *LemLat* adopts a rule-based approach which first splits the token into three parts in order to perform morphological tagging, namely the invariable part of the wordform (LES, e.g. *antiqu-*), the paradigmatic suffix (SM, e.g. *–issim-*) and the ending (SF, e.g. -*orum*) [Passarotti, 2007]. Like *LASLA*, *LemLat* is unable to contextually disambiguate ambiguous forms in running text, since it is lexicon-based, and more specifically makes use of the dictionaries Georges, Gradenwitz and the Oxford Latin Dictionary.

*1.2.6 LatinISE*

The *LatinISE* corpus comprises a total of ±13,000,000 Latin words covering a time span from the 2nd century B.C. to the 21st century A.D., was annotated through a combination of two pre-existing methods. Firstly, the *PROEIL* Project's morphological analyser and Quick Latin were used for lemmatization and PoS-tagging [McGillivray and Kilgariff, 2013]. This analyser generated various options in disambiguation for a word. Secondly, the output from the analyser was the input to a TreeTagger model trained on the *Index Thomisticus* dataset, which would take context into account and choose the most likely lemma and PoS-tag. *LatinISE* can be accessed online on the Sketch Engine, but is not freely available.

*1.2.7 CompHistSem*

A more recent promising project is *CompHistSem* (*Computational Historical Semantics*). The team has applied network theory to detect semantic changes in diachronic Latin corpora. Recently, they have released a composite lexicon called the *Frankfurt Latin Lexicon,* also referred to as the *Collex.LA*, which brings together lemmas from various web-based resources (such as the Latin Wiktionary) [Mehler et al, 2014].[3] Additionally the *TTLab Latin Tagger* was released, which has the objective to automatically tag large corpora such as the *Patrologia Latina*. Both these resources, *Collex.LA* and the *TTLab Latin Tagger*, are available for trial online. Their *TTLab Latin Tagger* is hybrid, in that it combines a linguistic rule-based approach with a statistical one, avoiding the huge effort rule-based taggers require for every target language separately on the one hand, and avoiding the overfitting characteristic of statistical taggers on the other [Mehler et al, 2014]. They have trained and tested the *TTLab Latin Tagger* on the Carolingian *Capitularia* —i.e. ordinances in Latin decreed by Carolingian rulers, split up in several sections or chapters.

[3] "AGFL Grammar Work Lab, the Latin morphological analyzer LemLat, the Perseus Digital Library, William Whitaker's Words, the Index Thomisticum (sic), Ramminger's Neulateinische Wortliste, the Latin Wiktionary, Latin training data of the Treetagger, the Najock Thesaurus [...] and several other resources. Beyond that, the FLL is continuously manually checked, corrected and updated by historians and other researchers from the humanities" (90). In the meantime, they report to have 8.347.062 word forms, 119.595 lemmas and 104.905

In a recent publication, the team has contributed to the field with an oversightful survey paper in which they have employed these *Capitularia* as training data to produce a comparative study of six taggers and two lemmatization methods [Eger et al, 2015]. These results arguably offer the best discussion of the state of the art at present. Out of the six taggers which they compared, more specifically *TreeTagger*, *TnT*, *Lapos*, *Mate*, *OpenNLPTagger* and *Stanford Tagger*, the best tagger was reported to be *Lapos*. When it comes to lemmatization, the team concluded that a trained lemmatizer (as opposed to a lexicon-based lemmatizers) provides better results (from 93-94% to 94-95%) and moreover deals better with lemmatizing words which are out-of-vocabulary (OOV) or which suffer from several variations (*honos* and *honor*) [Eger et al, 2015]. They used *LemmaGen* for this specific purpose, which is a lemmatizer dependent on induced rule conditions (RDR, ripple-down-rules) [Juršič et al, 2010]. This proves that lexicon-based approaches to lemmatization are not always favourable.

In an upcoming article [Eger et al, forthcoming], they have further developed this idea, by showing that the lemmatizer *LAT*, which relies on statistical inference and treats lemmatization as a sequence labeling problem (involving context), provides better results than *LemmaGen*. Both lemmatizers were based on prefix and suffix transformations. Moreover, *CompHistSem* has shown how the 'joint learning' of a lemmatizer with a tagger (as opposed to 'pipeline learning', which is the independent training and testing on each subcategory as PoS, case, gender etc.) can also improve the overall accuracy of the lemmatization / PoS-tagging task, especially in the case of the *MarMoT* tagger which —once additional resources such as word embeddings[4] and an underlying lexicon such as *Collex.LA* are provided— gains the highest results. The *CompHistSem*-team was generous to provide us with the annotated *Capitularia*-corpus (and their exact train-test splits), which facilitates the comparison of our results to theirs.

### 1.3. General trends, remaining problems

The preceding survey demonstrates that for lemmatization and part-of-speech tagging we dispose of the following annotated data: the *Index Thomisticus Treebank*, the *Latin Dependency Treebank*, the *PROIEL* data and the *Capitularia* corpus. All of these annotated corpora offer at least a lemma, coarse PoS tags and a fine-grained morphological analysis.

Interestingly, three trends appear from the state-of-the-art survey.

(1) Firstly, The automatic annotation of Latin texts has been moving away from semi-automated, rule-based approaches (e.g. *PROIEL*, *LASLA*, *CHLT LemLat*) to data-driven machine learning techniques (e.g. the *TreeTagger* in *LatinISE* and *CompHistSem*). In general, older approaches were strongly dependent on static lexica, which for each word form would exhaustively list all potential morphological analysis, e.g. in the form of tag-lemma pairs. In the case of ambiguity, a statistically trained part-of-speech tagger would be used to later single out the best option. First of all, such a superlemmas (these are lemmas which cover a certain word in its different varieties) in their *Collex.LA* [Eger et al, 2015].

[4] A technique which enables a representation of words as vectors, containing real numbers in a low-dimensional space which has a size dependent of the vocabulary size.

lexicon-based lemmatization approach has the disadvantage that it is in principle unable to correctly lemmatize out-of-vocabulary words, which are not covered in the available lexica. In the case of medieval texts, orthographic variation renders this problem even more acute. Moreover, lexicon-based strategies are very susceptible to the problem of error percolation: if the trained tagger predicts the wrong PoS tag, this renders it less likely that the tagger will be able to select the correct lemma. The *CompHistSem* project leads the way in this respect, showing that statistical lemmatization techniques offer an interesting, and perhaps even more robust alternative to traditional lexicon-based approaches.

(2) Secondly, *CompHistSem*'s latest results demonstrate that taggers which include distributed word representations (so-called "embeddings", see *infra*) are generally superior to previous approaches. This observation will prove relevant in the next section, since our architecture makes use of a similar representation strategy.

(3) Very few systems have attempted to learn the tasks of lemmatization and PoS-tagging in an integrated fashion.[5] Most systems continue to learn both tasks independently although some systems would make use of cascade taggers, where the output of e.g. the PoS-tagger would be subsequently fed as input to the lemmatizer. Nevertheless, previous research has clearly demonstrated that both tasks might mutually inform each other [Toutanova et al, 2009].

## II AN INTEGRATED ARCHITECTURE

### 2.1. Introduction to architectural set-up

In this section, we describe our attempt at an *integrated* architecture which can be used for the automated sequence tagging of medieval Latin at several levels, e.g. combined lemmatization and PoS-tagging. This architecture is comparable in nature to other sequence taggers, such as *Morfette* [Chrupała et al, 2008; Chrupała 2008]. Our architecture is in principle language-independent and could easily be applied to other languages and corpora. While this section is restricted to a complete, but high-level description, minor details of the architecture and training procedure can be consulted in the code repository which is associated with this paper.[6] A graphical depiction of our complete model is depicted in Fig. 1. The overall idea behind the architecture is simple. We first create two 'subnets' that act as encoders: one subnet is used to model a particular focus token —the token which we like to tag—, and a second subnet serves to model the lexical context surrounding the focus token. The result of these two 'encoding' subnets is joined into a single representation which is then fed to two other 'decoding' networks: one which will generate the lemma, and another which will predict the PoS tag.

### 2.2 One-hot word representation

Latin is a highly inflected language. In order to arrive at a good model for individual words, it is vital to take into account morphemic information at the subword level. We make use of recent advances in the field of "deep" representation learning [LeCun et al, 2015; Bengio et al, 2013], where it has been demonstrated recently that (even longer) pieces of text can be efficiently modeled from the raw character level upwards [Chrupała 2014; Zhang et al, 2015; Bagnall 2015; Kim et al, 2016]. We therefore present individual words to the network using a simple matrix representation as follows: each row represents a character, and each column represents the respective character positions in the word (cf. [Zhang et al, 2015]). We set the number of columns to be the length of the longest word in the training material: longer words at test time are truncated to this fixed length, and shorter words are padded with all-zero columns. The cells are populated with binary 'one hot' values, indicating the presence or absence of a character in a specific position in the word. A simplified example of this representation is offered below in Table 1 (for the word *aliquis*). All tokens are lowercased before this conversion, in order to limit the size of the character vocabulary.

---

[5] Note that our notion of the integration of these tasks differs from what others have called the 'joint' learning of complex part-of-speech tags.

[6] https://github.com/mikekestemont/pandora

| Char/position | 1 (a) | 2 (l) | 3 (i) | 4 (q) | 5 (u) | 6 (i) | 7 (s) | 8 (-) | 9 (-) | 10 (-) |
|---|---|---|---|---|---|---|---|---|---|---|
| **a** | 1 | 0 | 0 | 0 | 0 | 0 | 0 | 0 | 0 | 0 |
| **l** | 0 | 1 | 0 | 0 | 0 | 0 | 0 | 0 | 0 | 0 |
| **i** | 0 | 0 | 1 | 0 | 0 | 1 | 0 | 0 | 0 | 0 |
| **q** | 0 | 0 | 0 | 1 | 0 | 0 | 0 | 0 | 0 | 0 |
| **u** | 0 | 0 | 0 | 0 | 1 | 0 | 0 | 0 | 0 | 0 |
| **s** | 0 | 0 | 0 | 0 | 0 | 0 | 1 | 0 | 0 | 0 |

**Table 1**: Example of the character-level representation of an individual focus token (*aliquis*): this representation encodes the presence of characters (one per row) in subsequent positions of the word (the columns). Shorter words are padded with all-zero columns.

## 2.3 Long Short-Term Memory (LSTM)

### *2.3.1 Advantages to LSTM*

Next, we model these matrix-representations of words using so-called 'Long Short-Term Memory' (LSTM) layers [Hochreiter et al, 1997; Graves et al, 2013]. LSTM is a powerful type of sequence modeler to which is currently paid a great deal of attention in the field of representation learning, in the context of natural language processing in particular [LeCun et al, 2015]. This sort of 'recurrent' modeler will iteratively work its way through the subsequent positions in a time series, such as the character positions in our matrix. At the end of the series, the LSTM is able to output a single, dense vector representation of the entire sequence. LSTMs are interesting sequence modelers, because they arguably capture longer-term dependencies between the information at different positions in the time series. From the point of view of the present task, one could for instance expect the LSTM to develop a sensitivity to the presence of specific morphemes in words, such as word stems or inflectional endings. LSTM layers can be stacked on top of each other, to obtain deeper levels of abstraction. In our experiments, we use stacks of two LSTM layers throughout.

### *2.3.2 Word embeddings*

Apart from this character-level representation, our network architecture has a separate subnet which we use to model the lexical neighbourhood surrounding a focus token for the purpose of contextual disambiguation. We used the series of tokens starting from two words before the focus token until (and including) the token following the focus token (including the focus token itself), which is common contextual parametrization in this sort of sequence tagging [cf. Zavrel et al, 1999]. This part of the network is based on the concept of so-called 'word embeddings' [Baroni et al, 2010; Mikolov et al, 2013; Manning, 2015]. In traditional machine learning approaches, words are represented using their index in a vocabulary: in the case of a vocabulary consisting of 10,000 words, each token would get represented by a vector of 10,000 binary values, one of which would be set to 1, and all others to zero (hence, the alternative name 'one-hot encoding'). Such a representation has the disadvantage that it requires word vectors of a considerable dimensionality (e.g. 10,000). Moreover, it is a categorical word representation that judges all words to equidistant, which is of course a less useful approximation in the case of synonyms or spelling variants.

In the case of 'embeddings', tokens are represented by vectors of a much lower dimensionality (e.g. 150), which offer word representations in which the available information is distributed much more evenly over the available units. The general idea is that such word embeddings do not only come at a much lower computational cost, but that they also offer a smoother representation of words, because they are able to reflect, for instance, the closer semantic distance between synonyms. From a computational perspective, learning word embeddings typically involves optimizing a randomly initialized matrix [Levy et al, 2015], which for each vocabulary item holds a fixed-size vector of e.g. 150 dimensions. While such a matrix can quickly grow very large, word embeddings are still very efficient, because for each token only a single vector in the matrix has to be updated each time, leaving the rest of the matrix unaltered. In our network architecture, modelling the context surrounding a focus token as such involves to select 4 vectors from our embeddings matrix, and concatenate these into a single vector.[7] This yields a model of the focus token, as well as the surrounding context. We can consequently concatenate both 'encoding' representations into a single 'hidden representation'. The decoding parts of our network will produce the ultimate output for a focus token.

### *2.3.3 Training an LSTM*

As to the lemmatization, we now feed our 'hidden representation' into a second stack of LSTMs, by repeating this representation *n* times, where *n* corresponds to the maximum lemma length encountered during training. The task of this 'decoding' LSTM is to produce the correct lemma by generating the required lemma *character by character*. This is an extremely challenging approach to the problem of lemmatization. Lemmatization was previously approached in a more conventional classification setting: either the lemma was considered an atomic class label [Kestemont et al, 2010] or the lemmatization was solved by predicting an 'edit script' as a class label [Chrupała 2008; Chrupała et al, 2008; Eger et al, 2015], which could be used to convert the input token to its lemma. Instead of having the LSTM-stack output a single vector —as was the case for the encoder—, we now output a probability distribution over the characters in our alphabet for each character slot in the lemma. Therefore, we represent the lemma as a character matrix, using the exact same representation method as for the input tokens (cf. Table 1). We borrow the idea of an encoder-decoder LSTM architecture from a seminal paper in the field of Machine Translation which showed that stacks of encoding/decoding LSTMs can be used to transduce sentences from a source language into a target language [Sutskever et al, 2015; Cho et al, 2014]. Here, however, we do not learn to map a series of words in one language to a series of words in another language, but we use it to translate the series of characters in a token, to a series of characters representing the corresponding lemma.

The proposed network architecture is 'multi-headed' [Bagnall 2015], in the sense that a single architecture is used to simultaneously solve multiple tasks in an integrated fasion. Apart from the 'lemma-head', we also add a second 'head' to the architecture which aims to predict a PoS tag for a focus token. We use the 'hidden representation' obtained from the encoder and feed it into a stack of two standard dense layers [LeCun et al, 2015]. As is increasingly common in representation learning, we apply dropout to these layers ($p$=.5), meaning that during training, each time randomly half of the available values in a vector are set to zero [Srivastava, 2014]. As a non-linearity, we use rectified linear units, which set all negative values to zero. Finally, we produce a probability vector for each PoS label, which is normalized using a so-called softmax layer, ensuring that the resulting probabilities sum to one.

[7] In reality, this vector is then projected onto a more compressed layer, using a standard dense layer.

We train this network during a maximum of 15 epochs using an optimization method called 'minibatch gradient descent', with an initial batch size of 100. More specifically, we used the RMSprop update mechanism, which helps networks to converge faster because it keeps track of the recent gradient history of each parameter. After 10 epochs, we would decrease the current learning rate by a factor of three and the initial batch size with a factor to get more fine-updates for each batch. Some more specific implementation details: we do not apply any dropout in the recurrent layer as it proved to be detrimental; the recurrent layers use the *tanh*-nonlinearity, and all other nonlinearities which we tested failed to converge. All recurrent layers and dense layers have a dimensionality of 150, with the exception of the final output layer of the encoding LSTM, which we set to 450. We used a dimensionality of 100 for the embeddings matrix (below, we offer a visualization of one of its optimized versions). We implemented these models using the *keras, sklearn, gensim* and *theano* [Bastien et al, 2012] libraries and trained them on an NVIDIA Titan X. Depending on the model's complexity and current batch size, one epoch on average would take between 600 and 1000 seconds.

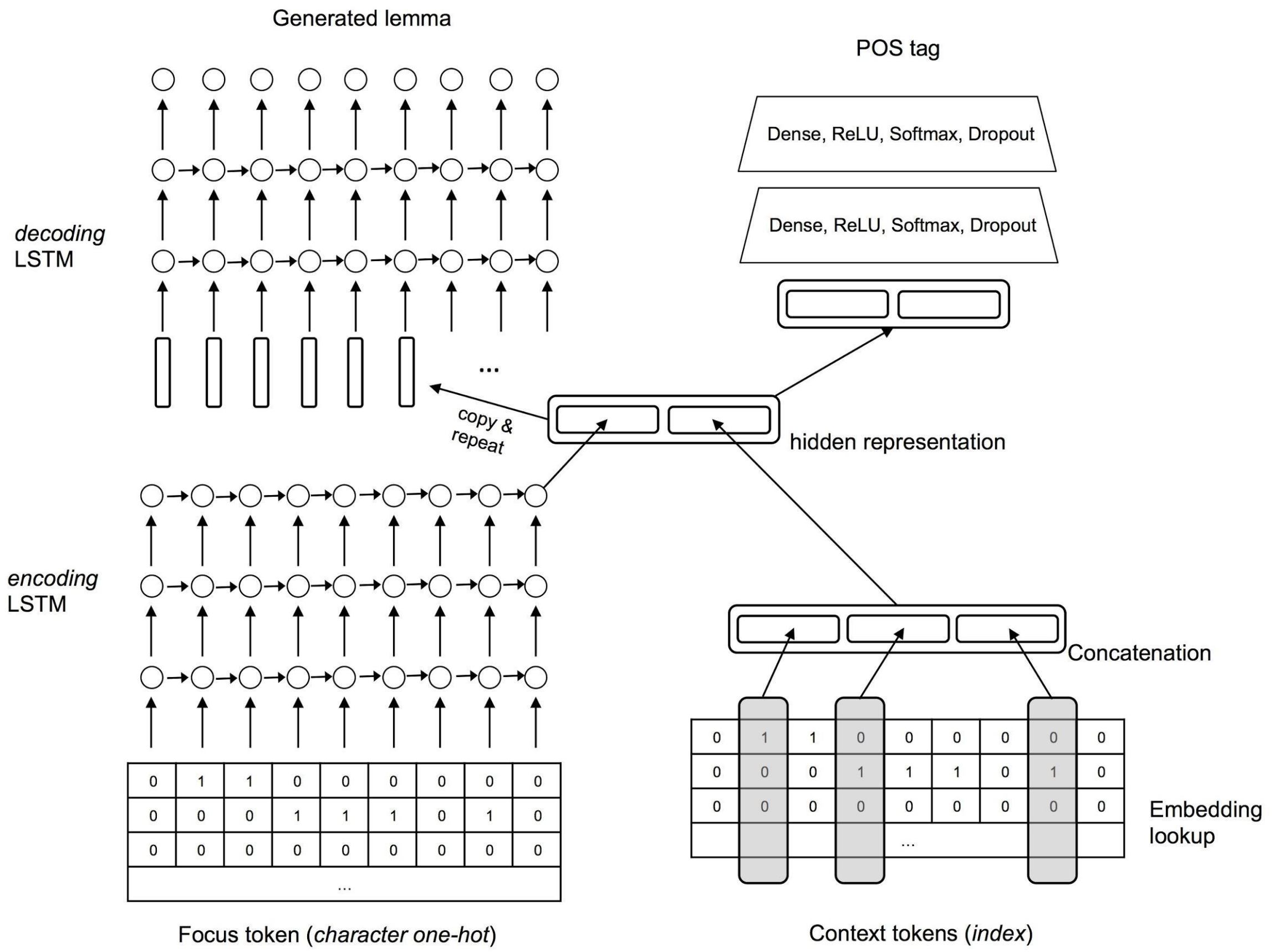

**Figure 1** Graphical representation of the proposed model architecture. The model has two 'encoding' subnets, which model a focus token and the surrounding context: the result is concatened into a single hidden representation. This represent is fed to two 'headnets': one which aims to generate the target lemma on a character-by-character basis; a second which predicts the PoS tag.

## III DATASETS AND EVALUATION METRICS

| | Non-Classicized | | | Classicized | | |
|---|---|---|---|---|---|---|
| | **Train** | **Dev** | **Test** | **Train** | **Dev** | **Test** |
| **Tokens** | 389,304 | 43,256 | 49,018 | 389,304 | 43,256 | 49,018 |
| **Unique tokens** | 38,044 | 10,353 | 11,526 | 38,045 | 10,353 | 5.71 |
| **Prop. unseen** | NA | 5.14 | 5.71 | NA | 5.14 | 5.71 |
| **Unique lemmas** | 12,413 | 4,904 | 5,256 | **10,906** | **4,568** | **4,837** |

**Table 2**: Statistics on the two datasets used in terms of number of words etc: the test set in the non-classisized spelling is identical to the one used by Eger et al. [2015].

### 3.1 Evaluation specifications

In this paper we evaluate the performance of our models using the traditional accuracy score (i.e. the ratio of correct answers over all answers). As is common in linguistic sequence tagging studies, we make a distinction between known and unknown tokens in the development and test data. "Unknown" tokens refer to the predictions for surface tokens which were not verbatimly encountered in the training data (which does not say anything about whether the target lemma for that surface form was encountered during training or not). In the training of neural networks, it has become standard to differentiate between a training set, a development set and a test set. The general idea here is that algorithms can be trained on the training data during a number of iterations: after each epoch, the system will gain in performance and can be evaluated on the held-out development data. When the performance of the system on the development data is no longer increasing, this is a sign that the system is overfitting the training data and will not generalize or scale well to the unseen data. At this point, one should halt the training procedure (a procedure also known as 'early stopping'). Finally, the system can be evaluating on the actual test data; this testing procedure should be postponed to the very end, to guarantee that researchers have not been optimizing a system in the light of a specific text set.

We have used the exact same test data as Eger et al. [2015], whose data set we will be focusing on. For development data, we have used the final 10% of instances of the remaining data; the first 90% were used as training data. Importantly, while this is a very objective approach to evaluating our system, this division of the data will put our architecture at a slight disadvantage in comparison to previous studies, in the sense that our system will only have been trained on 90% of all the available training data. Thus, our models can be expected to have a slightly worse lexical coverage, which might result in slightly lower scores etc. One important aspect of PoS-tagging is the complexity or granularity of the tagset used, which has of course an important impact on the performance of a tagger. In this exploratory paper, we limit our experiments to the simple PoS tags in the dataset, which only distinguish very basic word classes (e.g. N for nouns, V for verbs, etc.).

### 3.2 Making an additional "classicized" annotation layer in the dataset

One important issue with the original annotation standard used for the *Capitularia* data can be illustrated using the following example: consider the spelling of the word *oracio*, which has shifted from the classical *oratio* as a result of the lenition of the /t/ in medieval times. The current lemmatization standard will map both tokens to two separate lemmas, whereas they might just as well be mapped to the same lemma. For many projects (e.g. semantic or literary analysis), we would like a lemmatizer to collapse both spellings and map both to the same "superlemma", preferably in a uniform, "classical" spelling (for the sake of simplicity). We have therefore produced an alternative version of the *Capitularia* corpus, where all the training data's lemmas were normalized towards the classicized orthography, conventionally found in reference dictionary by Lewis & Short [Lewis and Short, 1879].

Amongst many, some of the more important rules are that both <v> and <u> are retained in their respective distinctions as consonant and vowel (auus or avvs is normalized to avus), <j> disappears for <i> (*conjunx* is normalized to *coniunx*), the diphtong <ae> is corrected or recovered where this is necessary (*aecclesia* to *ecclesia*, but *demon* to *daemon*), <ti> is recovered where <ci> is inappropriate in classical spelling (*rationem* instead of *racionem*), assimilations - especially in the case of prepositional prefixes - are allowed (*collabor* instead of *conlabor*), etc. Since many of the tokens in the *Capitularia* data had not been normalized to a standard spelling, we had to manually correct all deviant lemmas to the Lewis & Short norm, thus creating a resource to train models with classical spellings and lemmas. Regarding lemmatization conventions, the predominant principle is that all words are converted to their base form, which is the nominative singular for nouns, the nominative masculine singular for pronouns, adjectives and ordinal numbers, and the first person singular for verbs. Some choices are perhaps worth mentioning. For instance, comparatives and superlatives have been redressed to their neutral base forms (e.g. *maior* to *magnus*), gerunds and participles to their 1st person singular verb form. Adverbs retained their original form. Below, we will also report results using this dataset, which can be considered easier in the sense that the set of output lemmas is smaller, (see Table 2 for an overview), but more difficult in the sense that the character transduction between tokens and lemmas potentially becomes more complicated in the corrected cases.

## IV RESULTS AND DISCUSSION

### 4.1 Scores

As to the lemmatization results, our test scores are generally lower than the most successful scores reported by Eder et al. [2015], with an overall drop around 1.5-2.% in overall accuracy on the test set. This was partially to be expected, given the fact that our training only represents 90% of theirs, and thus has a slightly worse lexical coverage. Also, the formulation of the lemmatization task as a character-per-character string generation task is more complex, and currently does not seem to outperform more conventional approaches, in particular that of dedicated tools such as *LemmaGen*. Interestingly, however, our model is not outperformed by the results which Eger et al. [2015] reported for lexicon-based approaches, indicating that a machine learning approach relaxes the overall need for large, corpus-external lexica. Surprisingly, the accuracy scores for all tasks remain relatively low for the training data too, and none of our models reached accuracies over 96% for a particular tagging task, indicating the relative difficulty of the modelling tasks under scrutiny. The results for the 'classicized lemmas' version of the data set are generally in the same ballpark as the non-classicized data. This is a valuable result, since the string transduction task does in fact become more complex (although the set of output lemmas does shrink).

In this respect, it is worth pointing out that the results for the PoS-tagging task are relatively high and mostly on par with the best corresponding results reported by Eder et al. [2015]. This is somewhat surprising, given the limited training data that was used, as well as the fact that the model is fairly generic and does not include any of the more task-specific bells and whistles which current PoS-taggers typically include. Modern PoS-taggers often implement the recently predicted PoS tags of previous words as an additional feature to help disambiguate the current focus token. We did not include such features in our model, because they are not trivial to implement using a mini-batch training method. Nevertheless, our results suggest that our network produces excellent tagging results for the PoS labels. In all likelihood, this is due to the inclusion of distributed word embeddings, which have advanced the state of the art across multiple NLP tasks [Manning, 2016].

Note that in our implementation, we would first ‘pretrain’ a conventional word embeddings model on the training data, using a popular implementation of *word2vec*’s skipgram algorithm [Mikolov et al, 2013]. The fact that this pretraining data set is much smaller in size than the one used by Eger et al. [2015], i.e. the whole *Patrologia Latin*a, did not seem to pose a serious disadvantage. We used the resulting embeddings matrix, which is a cheap method to speed up convergence. Importantly, our word embeddings are dynamic, and the corresponding weight matrix will in fact be optimized during the training process to optimize them even further in the light of a specific task. Arguably, this is why our word embeddings approach is still on par with the approach reported by Eger. et al. [2015] where the word embeddings are added as a static feature, although they are trained on a much larger dataset. This creates interesting perspectives for future research. Below, we include a visualization of the word embeddings after training, using the popular *t-SNE* (t-distributed stochastic neighbor embedding) algorithm [Van der Maaten et al, 2008]. As visibly demonstrated, the model seems to learn useful representations of high-level word classes —e.g. prepositions form a tight cluster in light blue (*ex*, *in*, *pro*, …)— but also collocational patterns (*nostro tempore*, in light green).

For the integrated learning experiment, the results are curiously mixed: interestingly, in some respects, the tasks do seem to mutually inform themselves. The PoS results, for instance, are higher in the case of the integrated approach, which suggests that the PoS-tagging is helped by the information which is being backpropagated by the lemmatization-specific components. Surprisingly, however, this is not the case for lemmatization scores, which are actually lower in the integrated experiments. This is especially true for the unknown word scores. We hypothesize that the successful lemmatization of unknown words makes use of the surplus capacity in the hidden representation, or the capacity which is not strictly needed to predict the known word lemmas. In the integrated architecture, the PoS-tagger will require more information from the hidden representation, putting pressure on this surplus capacity. This strongly suggests that both tasks are to some extent competing for resources in the network, and further research into the matter is required.

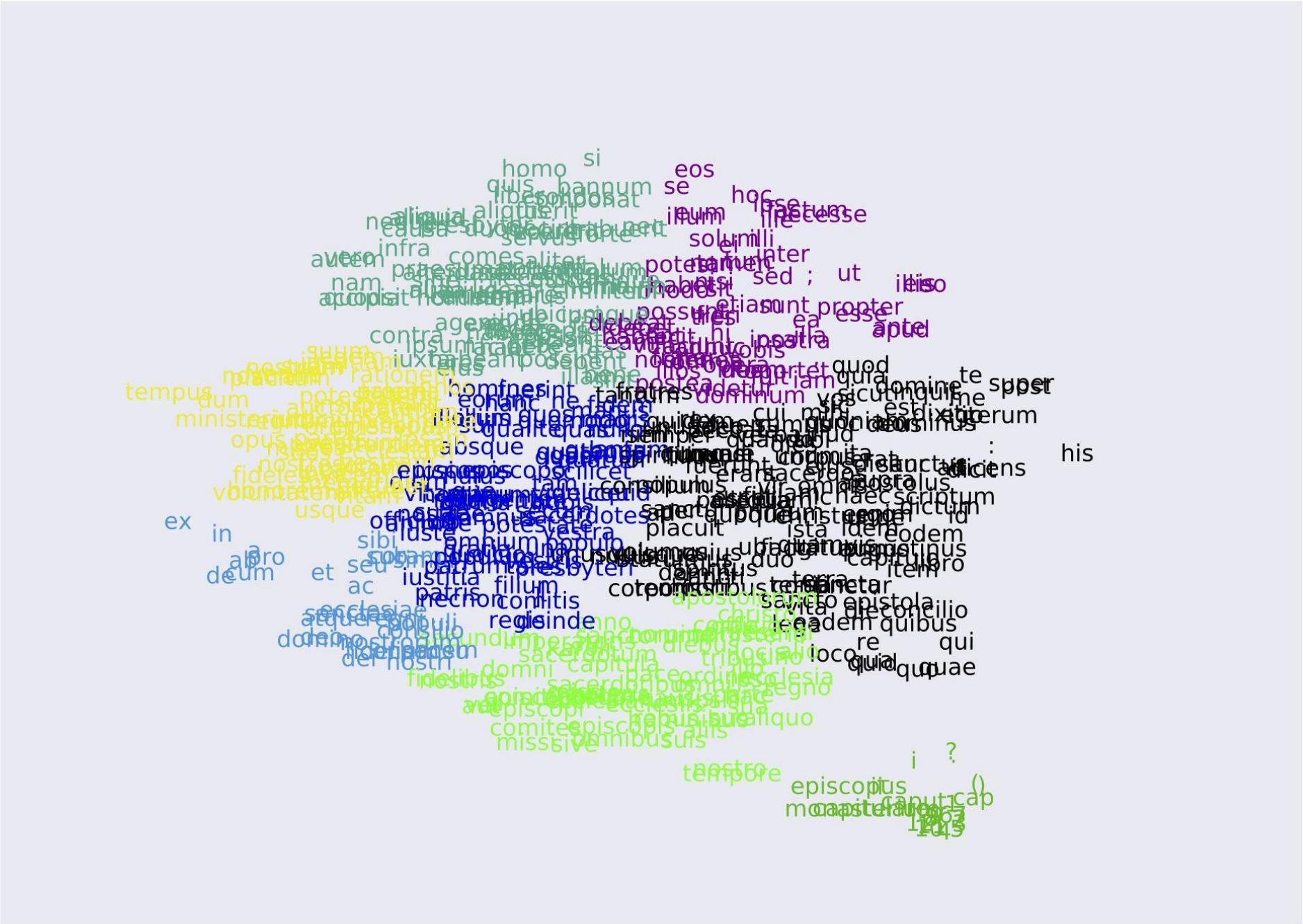

**Figure 2** A typical visualization of the word embeddings for the set of the 500 most frequent tokens in the training data after 15 epochs of optimization. A conventional agglomerative cluster analysis was ran on the data points in the scatterplot to identify 8 word clusters, which were coloured accordingly as a reading aid. These results are for the classicized corpus (integrated task of lemmatization and simple PoS tag prediction).

## 4.2. Output evaluation

As to an analysis of the errors outputted by the LSTM, one of the most recurrent was the intrusion of unwanted consonants or vowels. This is a predictable problem, since we generated the lemmas in a character-by-character fashion. Some false lemmatization results could perhaps best be described as a kind of "computational hypercorrection": the tagger attempts to solve a problem —i.e. set straight an orthographical variation— where it is in fact unnecessary. This is true for the normalization of *praesentaliter* to *praesintaliter*, in which a correction of the <e> to the <i> is observed, which we would only have expected with a token such as *quolebet*. Sometimes the tagger seemed sensitive to an orthographic problem but drew wrong conclusions in solving it, which is the case for *ymnus* being normalized to *omnis* instead of *hymnus*. Another typical problem is that proper names are not recognized as such, but as a different PoS, and are consequently "normalized" to an unrecognisable form. In general, we have noticed that this intrusion of consonants and vowels sometimes causes the fabrication of a lemma which is still quite far off from the lemma we wanted to predict, such as the lemmatization of *intromissi* to *intromittu*, or *lapidem* to *lapid*. Another interesting error was that *pluribus* was in some rare occasions lemmatized to *multus*, which indicates that the word embeddings in our model have struck a connection between two words that have semantic equivalence. This noise provides the pointers which we will need for solving this problem in future endeavours. Interestingly, a lot of the lemmatization errors which were eventually made, involve only small differences at the character level near the end of the lemma, which was in some ways to be expected since we generated the lemma left-to-right. Although some minor postprocessing might already be very helpful here, this suggests that the application of ‘bidirectional’ recurrent networks might be a valuable direction for future research [Graves et al, 2005].

| | Train | Dev | | | Test | | |
|---|---|---|---|---|---|---|---|
| **Task** | **All** | **All** | **Kno** | **Unk** | **All** | **Kno** | **Unk** |
| **Lemma** | 95.08 | 93.54 | 95.73 | 53.25 | 93.16 | 95.74 | 50.58 |
| **PoS** | 95.14 | 94.16 | 95.03 | 78.04 | 93.97 | 95.14 | 74.81 |
| **Lemma / PoS** | 92.09 / 95.59 | 91.03 / 94.50 | 93.53 / 95.34 | 44.85 / 78.98 | 90.54 / 94.44 | 93.57 / 40.37 | 40.37 / 75.63 |

**Table 3** Results (in accuracy) for the original, non-classicized lemmas in the *Capitularia* dataset [Eger et al, 2015]. Results are shown for the train, development and test set, for all words, as well as for the known and unknown words separately.

| | Train | Dev | | | Test | | |
|---|---|---|---|---|---|---|---|
| **Task** | **All** | **All** | **Kno** | **Unk** | **All** | **Kno** | **Unk** |
| **Lemma** | 95.40 | 93.91 | 96.22 | 51.27 | 93.57 | 96.27 | 48.91 |
| **Lemma / PoS** | 92.65 / 95.70 | 91.43 / 94.58 | 93.92 / 95.43 | 45.71 / 78.94 | 91.19 / 94.47 | 94.17 / 95.63 | 41.83 / 75.34 |

**Table 4** Results (in accuracy) for the *Capitularia* dataset with ‘classicized’ lemmas [Eger et al, 2015]. Results are shown for the train, development and test set, for all words, as well as for the known and unknown words separately.

## V CONCLUSION

This paper has presented an attempt to jointly learn two sequence tagging tasks for medieval Latin: lemmatization and PoS-tagging. These tasks are traditionally solved using a cascaded approach, which we bypassed by integrating both tasks in a single model. As a model, we have proposed a novel machine learning approach, based upon recent advances in deep representation learning using neural networks. When trained on both tasks separately, our model yields acceptable scores, which are on par with previously reported studies. Interestingly, our approach too is lexicon-independent, which places our results in line with previous studies with have moved away from lexicon-based approaches. When learned jointly, we observed that the PoS-tagging accuracy increased, but lemmatization accuracy decreased. Further research is required to discover how this competition for resources in the network can be handled in an efficient way. An important novelty of the paper is that we produced an annotation layer in the *Capitularia* dataset in which we normalized the medieval orthography of the lemma labels used by “classicizing” them. In spite of the increased difficulty of the string transduction task, our model performed reasonably well on this novel data in terms of lemmatization.

### Acknowledgements

Our sincerest gratitude goes out towards our colleagues Prof. Dr. Jeroen Deploige and Prof. Dr. Wim Verbaal from Ghent University, whose expertise and feedback was indispensable in the creation of this paper, which is a preliminary step in our joint project "Collaborative Authorship in Twelfth Century Latin Literature: A Stylometric Approach to Gender, Synergy and Authority", funded by the BOF research fund in Ghent. Without their considerate guidance in respectively the historical and linguistic / literary field, this article would been impossible. We would moreover like to express our gratitude to the *CompHistSem*-team that generously provided us with additional training data, and to the anonymous reviewer for his/her indispensable feedback in the process of revision.